# Contextual Semantic Parsing using Crowdsourced Spatial Descriptions

Kais Dukes (sckd@leeds.ac.uk)
Institute for Artificial Intelligence, University of Leeds
United Kingdom, LS2 9JT

## Abstract

We describe a contextual parser for the Robot Commands Treebank, a new crowdsourced resource. In contrast to previous semantic parsers that select the most-probable parse, we consider the different problem of parsing using additional situational context to disambiguate between different readings of a sentence. We show that multiple semantic analyses can be searched using dynamic programming via interaction with a spatial planner, to guide the parsing process. We are able to parse sentences in near linear-time by ruling out analyses early on that are incompatible with spatial context. We report a 34% upper bound on accuracy, as our planner correctly processes spatial context for 3,394 out of 10,000 sentences. However, our parser achieves a 96.53% exact-match score for parsing within the subset of sentences recognized by the planner, compared to 82.14% for a non-contextual parser.

## 1 Introduction

Semantic parsers are essential components of natural language (NL) understanding systems, with recent work focusing on both shallow methods such as semantic role labeling (Carreras and Màrquez, 2005; Palmer et al., 2010) and deep methods that directly parse natural language into complete representations (Zettlemoyer and Collins, 2007; Lu et al., 2008). We consider the different problem of using context to guide the parsing process. Our deep parsing task for robotic spatial commands is inspired by the rule-based SHRDLU (Winograd, 1972), a robotic arm that manipulates shapes on a board. In contrast, we adopt a data-driven approach by using a treebank annotated with a novel Linguistically Oriented

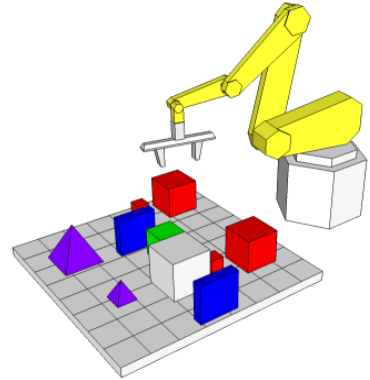

'*Pick up left purple prism and place on red cube one place in front of the one in the back right corner.*'

Figure 1: A complex spatial command with ellipsis ('place [it] on'), anaphoric references ('it' and 'one'), a multiword spatial expression ('in front of'), and lexical ambiguity ('one' and 'place').

Semantic Representation (LOSR), together with spatial scenes as additional context. Our task is challenging as multiword spatial expressions lead to attachment ambiguity (such as three different readings of 'move the red block on top of the blue cube on the yellow one'). Commands in the treebank are abbreviated so that ellipsis and anaphora are also common (Figure 1). However, our choice of representation makes semantic parsing more tractable. Instead of using a statistical model for lexical and attachment ambiguity, we use a *spatial planner*, a semantic component that determines if part of a LOSR description is compatible with a spatial scene.

In the next section we review the treebank, and in section 3 we survey related work. Section 4 describes a baseline experiment without spatial context. We present our contextual parser and its evaluation in sections 5 and 6 respectively.

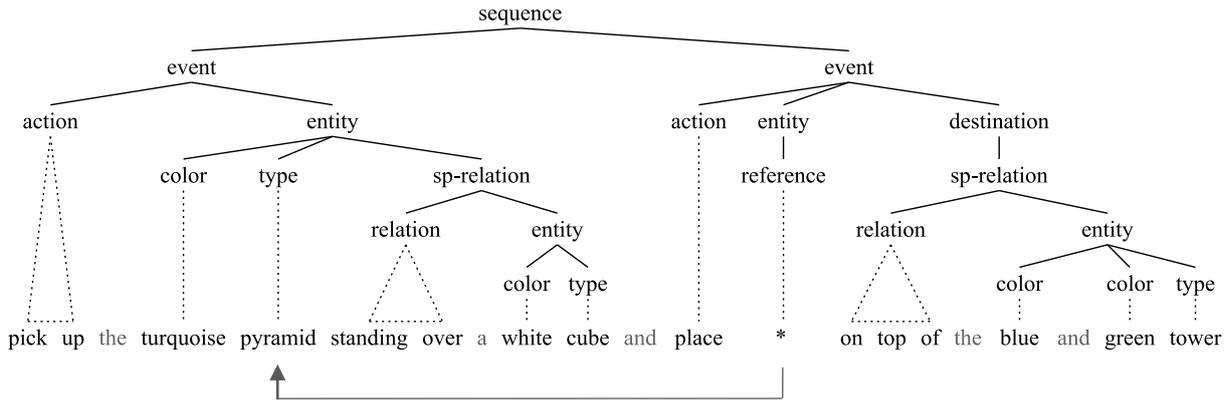

Figure 2: A word-aligned semantic tree with an elliptical anaphoric reference.

## 2 The Robot Commands Treebank

### 2.1 Annotation

Our dataset has 3,394 sentences (41,158 words) annotated out of 10,000 sentences collected via a new annotation game (Dukes, 2013), inspired by other games with a purpose such as Phrase Detectives (Chamberlain et al., 2009) and Google Image Labeler (Ahn and Dabbish, 2004). During data collection, participants are shown pairs of before and after images of scenes which are challenging to describe spatially, and are asked to give a command to a hypothetical robot to rearrange shapes from the first board configuration to the second. During offline annotation, we translated the collected sentences manually into LOSR. Sentences are only included in the treebank if the command specified by the corresponding LOSR description is verified by the spatial planner and results in a spatial configuration matching the second image for that scene (34% of all sentences). The treebank also includes word-aligned semantic trees that map words to complete LOSR descriptions (Figures 2 and 3).

### 2.2 Semantic representation

Because our semantic representation is general, we argue that our approach to parsing is applicable to other tasks. LOSR uses typed *entities* (labeled with semantic features) that are connected using *relations* and *events*. This universal formalism is not domain-specific, and is inspired by semantic frames (Fillmore and Baker, 2001), a practical representation used for natural language

```
(sequence:
 (event:
   (action: take)
   (entity:
     (id: 1)
     (color: cyan)
     (type: prism)
     (spatial-relation:
       (relation: above)
       (entity:
         (color: white)
         (type: cube)))))
 (event:
   (action: drop)
   (entity:
     (type: reference)
     (reference-id: 1))
   (destination:
     (spatial-relation:
       (relation: above)
       (entity:
         (color: blue)
         (color: green)
         (type: stack))))))
```

Figure 3. LOSR description with co-referencing.

understanding (Dzikovska, 2004; Dukes, 2009; UzZaman and Allen, 2010; Coyne et al., 2010).

As our approach is data-driven, parsing using a semantic component to aid disambiguation is not restricted to our chosen spatial task. With minimal modification to our representation, we expect to be able to annotate similar treebanks using LOSR for other domains. However, our approach crucially relies on a planner to guide the parsing process, and so could only be adapted to domains for a which a planner could conceivably exist. For example, nearly all robotic tasks such as such as navigation, object manipulation and task execution involve aspects of planning. NL question-answering interfaces to databases or knowledge stores are also good candidates for our approach, since parsing NL questions into LOSR within the context of a database schema or an ontology could be guided by a query planner.

## 2.3 LOSR features

In the remainder of this section we introduce notation that will be used to describe the semantic parser. In a LOSR description such as Figure 3, a preterminal node together with its child leaf node correspond to a feature-value pair (such as the feature *color* and the constant *blue*). We envisage high-level concepts in LOSR such as events, entities and relations to be general, while features associated with these concepts to be customized for a specific domain. An example of two entity features that are not domain specific are *id* and *reference-id*, which are used for co-referencing, such as for annotating anaphora and their antecedents. For a specific domain, we let $F$ denote the set of possible features, and for each feature $f \in F$ we use the notation $V(f)$ to denote the set of possible values for that feature. For example, for the robotic commands domain, $V(action)$ are the moves used to control the robotic arm, while $V(type)$ and $V(relation)$ are the entity and relation types understood by the spatial planner.

## 3 Previous related work

In deep semantic parsing, a translation function maps an NL sentence onto a formal meaning representation. Previous work can be broadly categorized into direct parsing that perform the translation process directly, and parsers that utilize additional situational context for disambiguation.

A standard dataset used to benchmark direct semantic parsers is the GeoQuery corpus (Wong and Mooney, 2007), consisting of 880 geography questions annotated with logical form. In contrast, the Robot Commands Treebank includes the positions of shapes in a scene as additional situational context. Parsing frameworks that have been successfully applied to GeoQuery include combinatorial categorial grammar (CCG) (Zettlemoyer and Collins 2007; Kwiatkowski et al. 2010), synchronous context-free grammar (SCFG) (Wong and Mooney, 2007; Li et al., 2013) and the generative model by Lu et al. (2008), who induce a translation function using a hybrid tree representation.

Also comparable to this paper, are systems that perform parsing jointly with *grounding*, the process of mapping natural language descriptions of entities in an environment to a semantic representation. Work in this direction includes Tellex et al. (2011), who develop a small corpus of natural language commands for a simulated fork lift robot that are parsed into Stanford dependencies (de Marneffe et al., 2006), with grounding performed using a factor graph. Similarly, Kim and Mooney (2012) perform joint parsing and grounding using a probabilistic context-free grammar (PCFG) over a corpus of robot navigation commands. The work in this paper contrasts with previous approaches by focusing on resolving attachment ambiguity. Whereas previous work has considered the mapping process from NL to a semantic representation by selecting the most-probable parse tree, we consider the different problem of performing this translation using additional situational context for disambiguation, using a linguistically-oriented representation.

## 4 Parsing without spatial context

As a baseline experiment, we have retrained the hybrid tree semantic parser by Lu et al. (2008) on our dataset, to establish a benchmark accuracy score for mapping from NL to LOSR without contextual disambiguation. We did not use gold-standard alignment data from the treebank for the benchmark. Instead, Lu's parser acquires its own lexical entries during training, initialized using IBM's alignment model 1 (Brown et al., 1993).

We use 100 EM iterations to train the unigram model described by Lu et al. (2008). Using 10-fold cross-validation on 3,394 sentences from the treebank, the total time taken was 1.4 hours. For direct parsing, the hybrid tree model achieved an accuracy score of 82.14%, averaged across each of the 10 folds. A strict metric is used to measure accuracy whereby a parse tree is considered correct only if it exactly matches the expected LOSR description in the treebank, and as a consequence is recognized correctly by the planner.

## 5 Contextual parsing

### 5.1 Methodology

In this section we describe a new parsing algorithm that exploits the structure of LOSR to integrate semantic context. We first provide an overview of our methodology followed by a detailed description of pre-processing (section 5.2), generalized shift-reduce parsing (sections 5.3 and 5.4) and post-processing steps (section 5.5).

**Step 1: Chunking and tagging**

During pre-processing, words are grouped into chunks. We tag each chunk with a LOSR feature $f \in F$. For example, we tag the chunk 'pick up' as an *action*, and 'to the left of' as a *relation*. Stop words, such as determiners outside of multiword expressions, are discarded. Only the highest scoring tag sequence is provided to the parser.

**Step 2: Generalized shift-reduce parsing**

Chunks are placed into a queue, which is incrementally read until empty. After reading a word, we perform a parallel reduction, guided by production rules and verified by the planner. If the top of the queue indicates ellipsis, an additional empty node is created. The result is a parse forest with trees that are syntactically correct according to a context-free grammar derived from training data, and with attachment decisions that are semantically grounded according to spatial context. For example, subtrees such as 'the green prism on the red cube' are only included only if this is compatible with the corresponding scene.

**Step 3: Anaphora resolution and ranking**

As a post-processing step, anaphora resolution is performed for each tree in the forest. LOSR actions are then verified by the planner and incompatible parses are discarded. If more than a single parse tree remains, these are ranked using a scoring function. Ranking helps resolve lexical ambiguities. For example, the word 'blue' is generally used to refer to blue shapes, but also to light blue (cyan) shapes. If a scene contains shapes with both these colors, the planner will consider each of these groundings valid. Lexical scoring is used to distinguish these parses probabilistically.

### 5.2 Semantic chunking

We interpret chunking as a sequence labeling problem, using the IOB2 representation (Sang, 2000). In the standard approach for noun phrases, POS tags are used to detect chunk boundaries which are labeled in a second step. In contrast, we perform chunking for untagged text directly using semantic labels. Let $f \in F$ be a semantic feature. Using the IOB2 representation, words that start and are in an *f*-chunk are tagged as B-*f* and I-*f* respectively, with outside words tagged as O. Figure 4 shows the tag sequence for the example sentence from Figure 1:

```
B-ACTION     pick
I-ACTION     up
B-INDICATOR  left
B-COLOR      purple
B-TYPE       prism
O            and
B-ACTION     place
B-RELATION   on
B-COLOR      red
B-TYPE       cube
B-CARDINAL   one
B-TYPE       place
B-RELATION   in
I-RELATION   front
I-RELATION   of
O            the
B-REFERENCE  one
B-RELATION   in
O            the
B-INDICATOR  back
B-INDICATOR  right
B-TYPE       corner
```

Figure 4. IOB2 chunking using semantic tags.

To train a chunker, the word-aligned semantic trees described in section 2 are used to construct IOB2 sequences as training data. In contrast to syntactic chunking for noun phrases, we assume that chunks in a LOSR treebank are small multiword expressions. Therefore, a second order Hidden Markov Model (HMM) can be used to predict the tag sequence $\{z_1, \ldots, z_n\}$, assuming that

$$P(z_1, \ldots, z_n) = \prod_{i=1}^{n+1} P(z_i | z_{i-1}, z_{i-2})$$

Here, $z_{-1,0}$ and $z_{n+1}$ are special start and stop symbols respectively.

Under this assumption, the sequence labeling problem is analogous to part-of-speech tagging. Our chunker is implemented in Java using the open source *jitar* HMM trigram tagger.[1] Once trained, the tagger will predict the IOB2 representation for a new sentence $\{w_1, ..., w_n\}$. Reversing the representation, O tags are used to discard stop words when these occur outside of chunks. The resulting $N$ chunks $\{c_1, ..., c_N\}$ have tags $\{f_1, ..., f_N\}$ that are feature labels, so that $f_i \in F$ ($1 \leq i \leq N$).

### 5.3 Phrase lexicon

Using training data from the treebank, we construct a phrase lexicon used for parsing. Given an *f*-chunk for a word sequence $\bar{w}$, a lexical function $L(\bar{w}, f) \subseteq V(f)$ maps the chunk to possible values for that feature. For example:

$L$ ( 'light blue', *color* ) = { *cyan* }
$L$ ( 'blue', *color* ) = { *blue*, *cyan* }
$L$ ( 'place', *type* ) = { *tile* }
$L$ ( 'place', *action* ) = { *move*, *drop* }
$L$ ( 'standing on top of', *relation* ) = { *above* }

Each value $v \in L(\bar{w}, f)$ is additionally paired with a weight $\omega(v)$. These are calculated using relative frequencies in training data, so that

$$\sum_{v \in L(\bar{w}, f)} \omega(v) = 1$$

### 5.4 Semantic parsing

We parse an NL sentence in the context of a spatial scene, represented by a world model $M$. A function $E(e, M)$ provided by the planner maps a LOSR entity description $e$ to a set of groundings in the world model. Similarly, the planner provides a predicate $A(a, M)$ which is true when a LOSR command $a$ is a valid action for a scene. In principle, parsing can then be performed through exhaustive search. Using a context-free grammar induced from training data, these two planning functions can be used to check if possible parses are compatible with spatial context.

---

[1] https://github.com/danieldk/jitar

In practice, we use dynamic programming to track previously verified LOSR descriptions. We use a graph-structured stack (GSS) for dynamic shift-reduce parsing (Tomita, 1988), an approach previously used for near linear-time dependency parsing (Huang and Sagae, 2010), efficient CCG parsing (Merity and Curran, 2011) and semantic disambiguation (Schiehlen, 1996).

```
1   Q = (c_1, ..., c_N)
2   G = ∅
3   R = ∅
4   C = ∅
5   while Q ≠ ∅ do
6       shift
7       reduce
8       if add-ellipsis then
9           reduce
10      end
11  end
```

Figure 5. Shift-reduce parsing loop.

Inspired by Merity and Curran (2011), we organize our GSS into *frontiers*, where each frontier is the list of vertices pushed onto the graph in one iteration of the main parsing loop. As a generalization of the stack used in deterministic shift-reduce parsing, paths in a GSS represent parallel stacks for different parsing choices. These stacks are kept synchronized through a shared shift operation. However, in contrast to previous approaches, we include an extra step to create elliptical nodes (Figure 5). Formally, the semantic parser's state is a tuple ($Q$, $R$, $G$, $C$) where:

1. $Q$ is an input queue.

2. $R$ is a reduction queue.

3. $G$ is a GSS, a directed acyclic graph where vertices represent shared (packed) LOSR subtrees.

4. $C$ are the vertices in the current frontier.

In the parser's initial configuration, the input queue holds chunks, with all other state empty. In the remainder of this section, we describe the shift, reduce and ellipsis operations.

**Shift:** Let $(Q, R, G, C)$ and $(Q', R, G', C')$ denote the parser's state before and after a shift operation respectively, so that

$$Q = (c_k, c_{k+1}, \ldots, c_N) \text{ and } Q' = (c_{k+1}, \ldots, c_N)$$

During a shift operation, we create a new frontier. Let $C$ denote the previous frontier with vertices $C = (\gamma_1, \gamma_2, \ldots)$. The chunk $c_k$ is a sequence $\overline{w}_k$ with feature tag $f_k$. Using the lexicon, for each possible value $v_i \in L(\overline{w}_k, f_k)$ we add a new GSS vertex $\xi_i$ to $G'$ holding a LOSR preterminal $f_k$ with leaf node $v_i$. Each vertex $\xi_i$ has a directed edge pointing to all vertices in the previous frontier $C$. The new frontier is then $C' = (\xi_1, \xi_2, \ldots)$. For example, Figure 6 shows the GSS after shifting $\overline{w}_k$ = 'place' with $f_k$ = *action*.

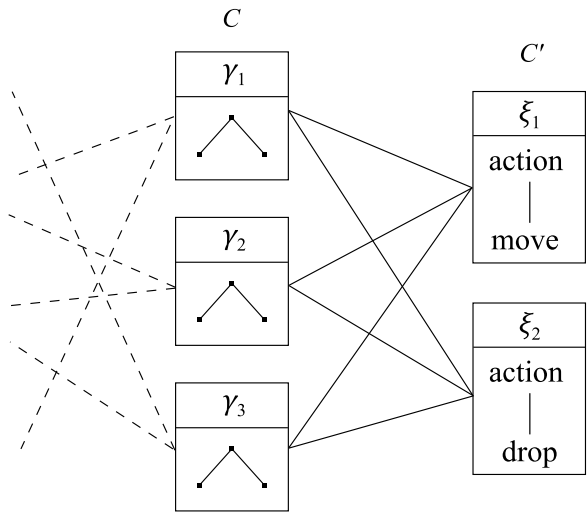

Figure 6. Shift operation with a new frontier.

**Reduce:** Let $(Q, R, G, C)$ and $(Q, R', G', C')$ be the state before and after the reduction stage. The queue $R'$ is initialized with vertices from the shift operation ($R' = C$). The following steps are repeated until the reduction queue is empty:

1. For each production rule, we search the GSS backwards from the current frontier.

2. For each path matching a rule, a new candidate vertex $\xi$ is constructed holding a parent LOSR node with child nodes from the vertices in the path $(\alpha_1, \ldots, \alpha_j)$.

3. If the node is a non-anaphoric entity $e$, we check if it is compatible with spatial context by determining if it has any groundings, i.e. if $|E(e, M)| \geq 1$.

4. If the node held by $\xi$ is a grounded entity (or is an anaphor or not an entity), it is added to the GSS, the current frontier, and to the reduction queue $R'$. We add directed edges from the new vertex $\xi$ to the vertices $(\beta_1, \beta_2, \ldots)$ further down the GSS connected to $\alpha_j$ (Figure 7).

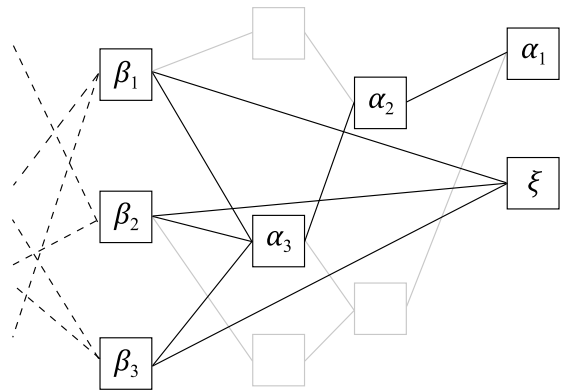

Figure 7. Non-destructive reduce operation.

For CGG parsing, Merity and Curran (2011) perform frontier pruning statistically. In contrast, we perform this semantically. In step 3 of reduction, an entity is only included if it is an anaphor, or if it is semantically grounded with realizations in the world model. As we show in our evaluation, this step allows the parser to perform in near linear-time by excluding invalid attachment decisions as soon as they arise.

**Ellipsis:** To trigger ellipsis, we use feature tags $f_k$ and $f_{k+1}$ of the top two chunks on the queue $Q = (c_k, c_{k+1}, \ldots, c_N)$. From training data, we build a table of rules that determine if an elliptical node should be added. An example rule would be to add an anaphoric elliptical node between $f_k$ = *action* and $f_{k+1}$ = *relation*, as in 'place [it] on'. Similar to a shift operation, if ellipsis is triggered we create a new frontier holding the elliptical vertex, followed by another reduce operation in the main loop (Figure 5).

## 5.5 Post-processing

**Anaphora resolution:** For the sentences in the Robot Commands Treebank, most anaphora are trivially resolved using pattern matching. For sentences of the form, 'pick up *X* and put it …', we resolve the anaphora 'it' to *X*. Otherwise, we resolve using the nearest preceding entity. For example, we resolve 'one' in 'put the red block on the yellow one' to 'the yellow *block*'.

**Lexical scoring:** The parser is sensitive to the lexical associations derived from training data. In the final step, each LOSR action tree $a$ in the forest is verified using the planner predicate $A(a, M)$. The remaining trees are lexically scored. Let $v(a) = \{v_1, …, v_m\}$ be the leaf feature-values for tree $a$. Under independence assumptions, we approximate the most probable tree by the tree with the best weights $\omega(v_i)$ (as per section 5.3):

$$P(v_1, …, v_m) = \prod_{i=1}^{m} P(v_i | v_{i-1}, … v_1)$$

$$= \prod_{i=1}^{m} P(v_i) = \prod_{i=1}^{m} \omega(v_i)$$

Therefore, the final tree $\hat{a}$ is chosen according to

$$\hat{a} = \underset{a}{\operatorname{argmax}} \prod_{x \in v(a)} \omega(x)$$

## 6 Evaluation

### 6.1 Performance

The treebank contains 3,394 sentences that have an average of 12.1 words. For evaluation, we use 10-fold cross-validation, measuring a parse tree as correct if it exactly matches the treebank. In its default configuration, the contextual parser scored 96.53%, compared to 82.14% for the non-contextual baseline (Table 1). To measure the effect of the different steps in our approach, we preformed three further experiments. Without lexical scoring, accuracy was 81.66%, as multiple parses (all considered compatible by the planner) could not be disambiguated. In compari-son accuracy was 88.78% by randomly selecting a parse that was compatible with spatial context. Finally, we considered the effect of the HMM tagger, by removing this from the pipeline and providing perfect chunks to the parser using gold evaluation data, giving a 97.24% upper bound.

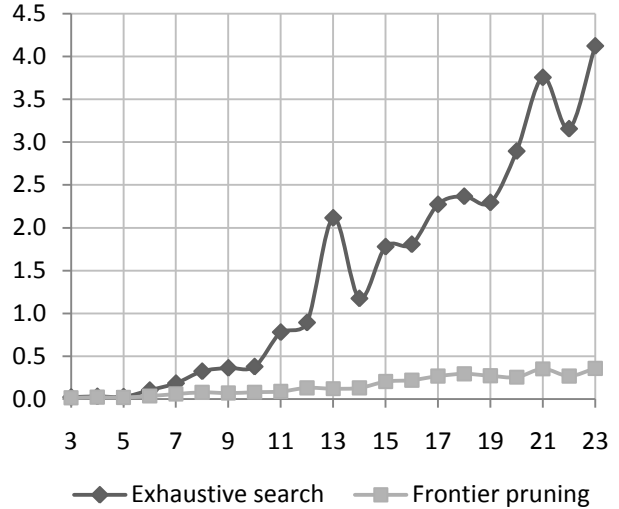

Figure 8. Average parsing time (in milliseconds) as a function of sentence length (word count).

| | |
|---|---|
| Non-contextual baseline (Lu et al.) | 82.14 |
| Contextual parser (without scoring) | 81.66 |
| Contextual parser (random selection) | 88.78 |
| Contextual parser (default) | **96.53** |
| Contextual parser (gold chunking) | 97.24 |

Table 1. Parsing accuracy using cross-validation.

Overall, the contextual parser was also faster than the baseline. Cross-validating 10 times, including training the tagger, extracting production and ellipsis rules, followed by evaluation (which included integrated parsing with spatial planning) took a total of 6.1 seconds. This compares to a total of 1.4 *hours* for the baseline. The difference is because the parser by Lu et al. (2008) uses an EM training process, although in contrast it is a more general model that is applicable to a wider range of direct parsing problems.

Figure 8 shows that frontier-pruning leads to approximate linear-time parsing. Although exhaustive search could be performed using CYK, GSS parsing does not require binarization and as we have shown, can also incorporate ellipsis. In comparison, GSS exhaustive search took a total of 19.76 seconds (3.2 times longer), with parsing accuracy the same as with frontier pruning.

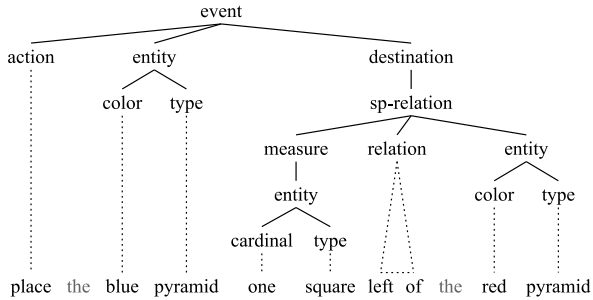

Figure 9. A quantitative relation ('one square left of').

### 6.2 Error analysis

Aggregated across all folds, 118 sentences were misparsed, out of 3,397 evaluated instances. The chunker contributed to 39 misparses. A further 45 were rejected due to out-of-vocabulary errors, as our algorithm fails when given unseen words. 16 errors were due to anaphora resolution. The remaining 18 errors were due to scoring, where the wrong tree was selected, or multiple parses were verified by the planner with the same score.

### 6.3 Discussion

Our parsing approach is dependent on the corresponding LOSR for an NL sentence being recognized by the planner. Through crowdsourcing we have collected 10,000 sentences, with 3,394 annotated into LOSR. This gives an overall 34% upper bound on accuracy using our current implementation of the planner. Our main contribution is that integrated parsing with planning gives an accurate result for this subset of sentences. The parser's performance for this subset can be attributed to two main factors. Firstly, the treebank has a small vocabulary of 600 words. It is known that English uses only around 70 spatial prepositions (Landau and Jackendoff, 1993; Herskovits, 1998), together with a small number of spatial expression types, such as quantitative relations (Figure 9). As such, sentences are not as linguistically diverse as other treebanks, simplifying tasks such as anaphora resolution. Secondly, we have chosen a linguistically-oriented representation that closely aligns with compositional sentence structure, streamlining integration.

## 7 Conclusion and future work

In this paper, we presented a new crowdsourced treebank of spatial descriptions, annotated using a novel linguistically-oriented semantic representation. We replaced the use of a statistical model for disambiguation by a semantic component for handling lexical and attachment ambiguity. We have also shown that this can be done efficiently, using a shift-reduce parser that runs in near linear time. A GSS with frontier pruning was used to rule out invalid attachment decisions early on in the parsing process, leading to a 3.2 times speed increase for our dataset, compared to exhaustive search. Our proposed solution also handles ellipsis. Although previous work has incorporated ellipsis into deterministic shift-reduce parsing (Dukes, 2013c), to the best of our knowledge this is the first work that incorporates ellipsis into a GSS shift-reduce parser.

In future work, we plan to generalize the planner. Although mapping NL descriptions to a formal spatial calculus is non-trivial (Kordjamshidi et al., 2010), we are improving the spatial planner to cover the remaining sentences. The next planned stage in our approach is grounded language acquisition (Krishnamurthy and Kollar, 2013), where we plan to jointly train the parser and planner to learn a semantic lexicon based on a small number of spatial primitives. This will allow us to process unknown words, which are currently not handled by our semantic parser. At present, to acquire a new vocabulary for a new domain, additional manual annotation is required, as our lexicon is extracted from an annotated treebank.

Our long term goal is to integrate planning with dialog processing and question answering. An argument often directed at the classic system SHRDLU, our inspiration for this work, is that it did not generalize well to other tasks (Dreyfus, 2009; Mitkov, 1999). In contrast, we propose a data-driven approach for semantic parsing with planning, using a new dataset that we hope will be of interest to the semantic parsing community.